\documentclass[10pt,twocolumn,letterpaper]{article}

\usepackage{iccv}              
\usepackage{bm, bbm}
\usepackage{amsmath, amssymb}
\usepackage{amsmath, amsfonts, amssymb}
\usepackage{xparse}
\usepackage{comment}
\usepackage{multirow}
\usepackage{makecell}

\usepackage[english]{babel}
\definecolor{iccvblue}{rgb}{0.21,0.49,0.74}
\usepackage[pagebackref,breaklinks,colorlinks,allcolors=iccvblue]{hyperref}

\def\paperID{3115} 
\def\confName{ICCV}
\def\confYear{2025}

\title{PCGS: Progressive Compression of 3D Gaussian Splatting}

\author{
  Yihang Chen \textsuperscript{1, 2}\thanks{Contribute equally.} \qquad Mengyao Li \textsuperscript{3, 2}\footnotemark[1] \qquad Qianyi Wu \textsuperscript{2}\thanks{Corresponding authors.} \qquad Weiyao Lin \textsuperscript{1}\footnotemark[2]\\ 
  Mehrtash Harandi \textsuperscript{2} \qquad Jianfei Cai \textsuperscript{2} \\
  \textsuperscript{1}Shanghai Jiao Tong University \qquad
  \textsuperscript{2}Monash University \qquad
  \textsuperscript{3}Shanghai University 
  \\
  {\tt\small \{yhchen.ee, wylin\}@sjtu.edu.cn, sdlmy@shu.edu.cn} \\
  {\tt\small \{qianyi.wu, mehrtash.harandi, jianfei.cai\}@monash.edu}
}

\begin{document}
\maketitle

\begin{abstract}
    3D Gaussian Splatting (3DGS) achieves impressive rendering fidelity and speed for novel view synthesis. However, its substantial data size poses a significant challenge for practical applications. While many compression techniques have been proposed, they fail to efficiently utilize existing bitstreams in on-demand applications due to their lack of progressivity, leading to a waste of resource.
    To address this issue, we propose \textbf{PCGS} (\textbf{P}rogressive \textbf{C}ompression of 3D \textbf{G}aussian \textbf{S}platting), which adaptively controls \textbf{both the quantity and quality} of Gaussians (or anchors) to enable effective progressivity for on-demand applications. Specifically, for quantity, we introduce a progressive masking strategy that incrementally incorporates new anchors while refining existing ones to enhance fidelity. For quality, we propose a progressive quantization approach that gradually reduces quantization step sizes to achieve finer modeling of Gaussian attributes. Furthermore, to compact the incremental bitstreams, we leverage existing quantization results to refine probability prediction, improving entropy coding efficiency across progressive levels.
    Overall, PCGS achieves progressivity while maintaining compression performance comparable to SoTA non-progressive methods. Code available at: github.com/YihangChen-ee/PCGS.
\end{abstract}

\vspace{-15pt}

\section{Introduction}
\begin{figure} [t]
    \vspace{-10pt}
    \centering   
    \includegraphics[width=1\linewidth]{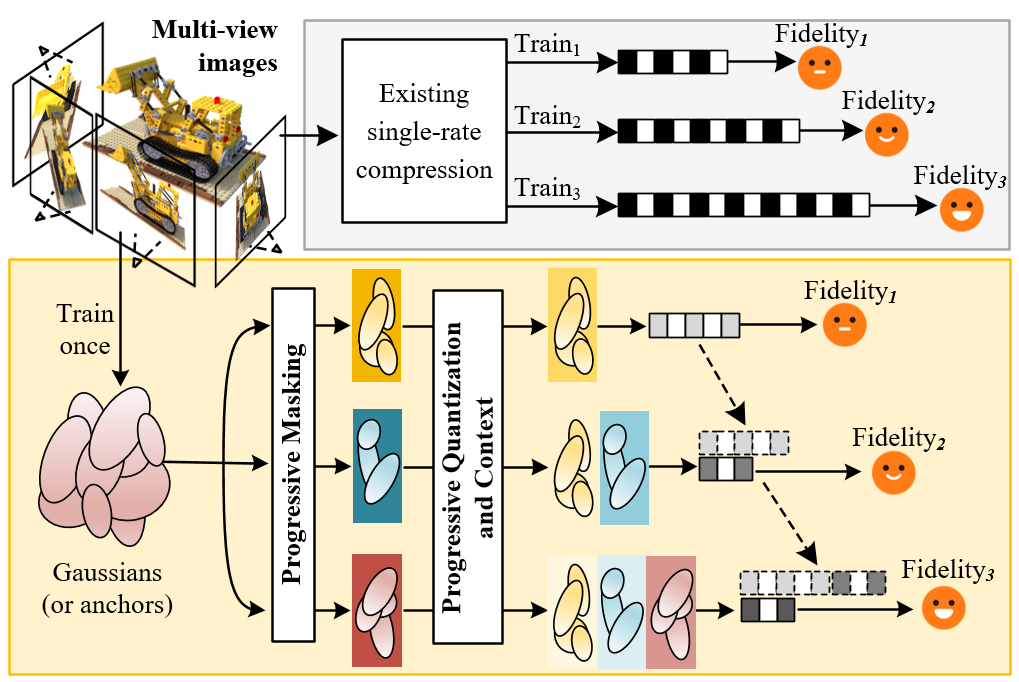}
    \caption{\textbf{Comparison of existing approaches (\textbf{upper}) and the proposed progressive compression (\textbf{lower}).} Existing approaches generate multiple independent bitstreams targeting different rates and fidelity through multiple trainings, while a progressive compression approach (with only one training) can continuously improve the fidelity by incrementally adding bitstreams, which is resource-saving in on-demand applications.}
    \vspace{-10pt}
    \label{fig:teaser}
\vspace{-10pt}
\end{figure}

In the field of novel view synthesis, 3D Gaussian Splatting (3DGS) \cite{3DGS} has emerged as a leading technology due to its photo-realistic rendering fidelity and real-time rendering speed. By explicitly defining Gaussians with color and geometric attributes, 3DGS provides a flexible and efficient 3D scene representation. However, the large number of Gaussians results in excessively large data sizes, making transmission and storage challenging \cite{BungeeNeRF, Trimming, 3dgszip}. To address this, various approaches have been proposed to compress 3DGS \cite{3dgszip, Lightgaussian, EAGLES, Trimming, SUNDAE, LP3DGS, SOG}. Despite these advancements, existing methods exhibit two key limitations: 
(1) Single-rate constraint: Each bitstream from these compression models is only for a fixed rate/data size; and (2) Lack of progressivity: Although these models can be retrained to produce bitstreams for different rate targets, the resulting bitstreams are independent and non-reusable, as illustrated in the upper part of Fig.~\ref{fig:teaser}.

These limitations become more pronounced in on-demand applications. For instance, as transmission bandwidth increases, bitstreams with higher rate budgets can be transmitted for better rendering fidelity. However, due to their single-rate and non-progressive nature of existing 3DGS compression methods, generating a new bitstream requires retraining the model, where existing bitstreams cannot be reused, leading to inefficient resource utilization.  
These inefficiencies are further exacerbated in large-scale 3DGS scenes \cite{mip360}, which often exceed $20$ megabytes (MB) even after compression. To overcome these challenges, progressive 3DGS compression presents a promising direction. The key idea is that \textit{existing bitstreams that encode base-quality 3DGS can be refined and augmented with additional bitstreams to achieve higher fidelity}, as illustrated in the lower part of Fig.~\ref{fig:teaser}.

Among existing 3DGS compression approaches, only a few studies have explored progressive 3DGS representations~\cite{lapisgs, gode,GaussianPro,PRoGS}. Specifically, LapisGS~\cite{lapisgs} constructs a layered structure of cumulative Gaussians to achieve progressively higher rendering resolutions, while GoDe~\cite{gode} organizes Gaussians into hierarchical layers to enable progressive compression. However, they primarily focus on increasing the quantity of Gaussians at each level while overlooking improving their quality.  
Moreover, they fail to exploit the inherent context across levels, which could be leveraged to reuse information from previous levels for enhanced compression efficiency. On the other hand, although context modeling has demonstrated its effectiveness in existing 3DGS compression methods~\cite{HAC, HAC++, CAT3DGS, Contextgs}, these approaches do not support progressive compression.
For instance, HAC++~\cite{HAC++, HAC} learns a sparse hash grid to capture contextual relationships among anchors, which are introduced in Scaffold-GS~\cite{scaffold} to predict nearby 3D Gaussians. By leveraging the context, HAC++ predicts the quantization step and the distributions of anchor attributes to facilitate entropy coding~\cite{balle2018variational}, significantly reducing the bitstream size.
Similarly, CAT-3DGS~\cite{CAT3DGS} employs multi-scale triplanes to model correlations among anchors. Despite the advancements, these methods remain single-rate, requiring separate trainings and generating independent bitstreams for different rate-distortion (R-D) trade-offs.

To address this, in this paper, we propose \textbf{PCGS}, a novel approach for the \textbf{P}rogressive \textbf{C}ompression of 3D \textbf{G}aussian \textbf{S}platting. \textbf{\textit{The core idea}} behind PCGS is to jointly and adaptively control both the quantity and quality of Gaussians (or anchors), enabling a progressive bitstream that incrementally enhances fidelity.
Specifically, to regulate the quantity of anchors, we introduce a monotonically non-decreasing mask learning strategy, which progressively decodes additional anchors and Gaussians, which are seamlessly integrated with existing ones at the decoder side to improve fidelity.  
For anchor quality, we progressively refine existing anchors by employing trit-plane division~\cite{trit}, which provides more precise quantization results.  
Building upon this refinement, we further introduce a quantization context model that exploits correlations among quantized values across progressive levels. It leverages a trinomial distribution to more accurately estimate the probability of quantized values, thereby improving entropy coding efficiency and reducing the incremental bitstream size.

Our main contributions are summarized as follows.
\begin{itemize}
    \item We propose \textbf{PCGS}, a novel progressive 3DGS compression framework that jointly and progressively enhances anchor quantity and quality. This enables diverse on-demand applications, significantly expanding the applicability of 3DGS.
    \item We introduce a \textit{progressive masking strategy} that incrementally decodes additional anchors and Gaussians and integrates them with existing ones to enhance fidelity. Furthermore, we propose a \textit{progressive quantization mechanism} that leverages quantization context to refine anchor quality and improve probability estimation.
    \item Extensive experiments across multiple datasets demonstrate that \textbf{PCGS} achieves effective progressivity while maintaining compression performance comparable to SoTA single-rate compression methods.
\end{itemize}

\section{Related Work}

\textbf{3D Gaussian Splatting.} 
3DGS represents a scene using a collection of 3D Gaussians \cite{3DGS}, each defined by learnable shape and appearance attributes. Leveraging the rasterization technique \cite{ewa}, these 3D Gaussians can be quickly trained and rendered into 2D images given a viewpoint.
While achieving real-time rendering speed and high fidelity, 3DGS typically demands significant storage \cite{3dgszip} and transmission bandwidth due to the large number of Gaussians and their associated attributes.

\noindent\textbf{3DGS Compression}.
Many efforts~\cite{3dgszip} have been made to reduce the data size of 3DGS while maintaining rendering fidelity. Some methods focus on reducing Gaussians parameters or compromising their precisions. Pruning techniques have been extensively explored, typically eliminating unimportant Gaussians through trainable masks \cite{Lee, LeeDynamic}, gradient-informed thresholds \cite{Trimming, ELMGS}, view-dependent metrics \cite{Lightgaussian}, and other importance-evaluation mechanisms \cite{LP3DGS, PUP3DGS}. 
Additionally, \cite{SUNDAE} and \cite{mini} propose to combine pruning with structural relations.
Instead of pruning entire Gaussian, \cite{SOG,papantonakis} partially prunes Gaussian attributes, such as Sphere Harmonics (SH) coefficients. 
Additionally, vector quantization \cite{Compact3d, SizeGS, Lee, Lightgaussian, Simon} groups similar Gaussians and represents them with a shared approximation. \cite{RDOGaussian} employs entropy-constrained vector quantization with codebooks to quantize covariance and color parameters, achieving a more compact representation.

Other methods \cite{IGS, scaffold, HAC, HAC++, Contextgs, CAT3DGS, Seungjoo2025} aim to exploit spatial structure correlations to reduce the size of 3DGS. \cite{IGS} predicts attributes of unstructured 3D locations using a multi-level grid for compact 3D modeling. Scaffold-GS \cite{scaffold} employs anchors to cluster region-near Gaussians and predicts the attributes of Gaussians within each cluster using an MLP. Building on Scaffold-GS, HAC \cite{HAC} and HAC++~\cite{HAC++} introduce hash-grid to explore the mutual context information between the attributes of anchors and hash features, facilitating entropy coding for a highly efficient representation. Different from HAC, CAT-3DGS \cite{CAT3DGS} employs multi-scale triplanes to capture spatial correlations.
HEMGS~\cite{HEMGS} proposes a hybrid entropy model for 3DGS entropy coding. ContextGS \cite{Contextgs} and CompGS \cite{liu2024compgs} reduce redundancy among anchors/Gaussians through context-aware designs.
GaussianForest \cite{GaussianForest} constructs hybrid 3D Gaussians hierarchically, where implicit attributes are shared among sibling Gaussians to save storage.
Additionally, optimization-free methods \cite{FCGS, MesonGS} are also explored to compress existing 3DGS rapidly in a single feed-forward pass. Overall, although these methods have achieved impressive R-D performance, they require multiple trainings to achieve different compression levels, incapable of producing progressive bitstreams, which are especially crucial in practice scenarios involving diverse bandwidth and device resource constraints.

\begin{figure*}[t]  
    \centering    
    \includegraphics[width=1.0\linewidth]{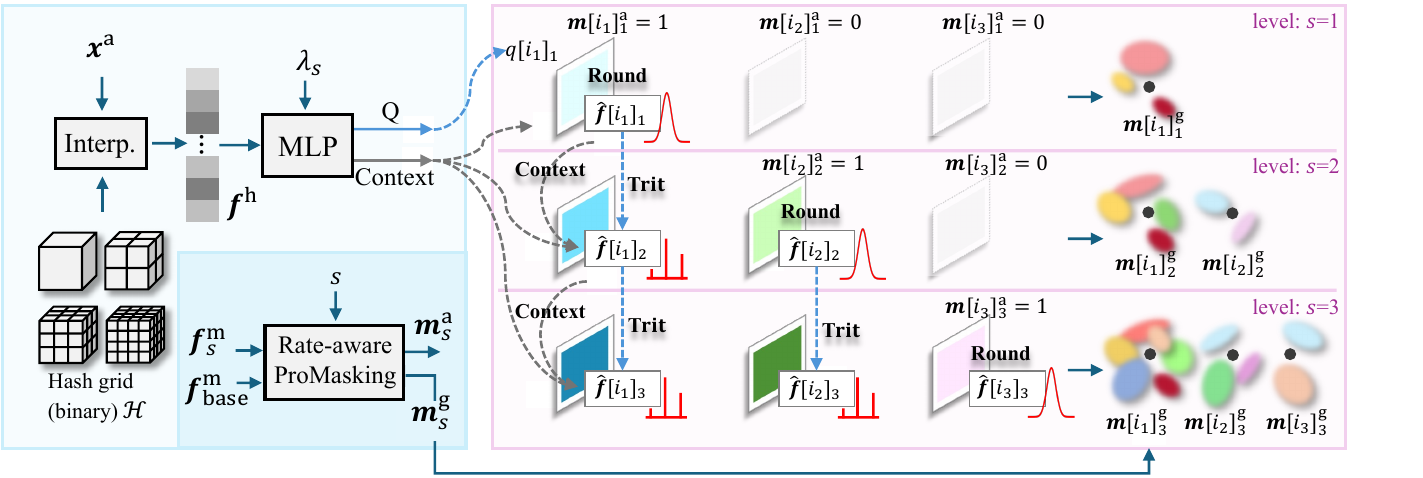}
    \vspace{-15pt}
    \caption{
    \textbf{Overview of the proposed PCGS}, which controls anchors in both quantity and quality in a progressive way, \ie, progressively \textit{decoding new anchors via masking control} and \textit{refining existing anchors with finer quantization steps}.
    \textbf{Left}: Given anchor $\bm{x}^\text{a}$, its position is interpolated within the binary hash grid to obtain the hash feature $\bm{f}^\text{h}$. A rate-aware MLP, conditioned on the level information $\lambda_s$, utilizes $\bm{f}^\text{h}$ to determine the quantization steps and provide context information for different progressivity levels. Additionally, the anchor and Gaussian masks $\bm{m}^a_{s}$ and $\bm{m}^g_{s}$ are obtained via the rate-aware progressive masking strategy from learnable features $\bm{f}^\text{m}_{s}$ and $\bm{f}^\text{m}_{base}$ (\textbf{bottom}). 
    \textbf{Right}: 
    At each level $s$, according to its mask $\bm{m}[i]^\text{a}_s$, the $i$-th anchor either remains undecoded when $\bm{m}[i]^\text{a}_s=0$, be newly decoded when $\bm{m}[i]^\text{a}_s$ transits from $0$ to $1$, or otherwise be refined.
    }
    \label{fig:main_method}
    \vspace{-15pt}
\end{figure*}

\noindent\textbf{Progressive 3DGS.} To accommodate on-demand applications, progressive 3DGS has been explored~\cite{lapisgs,He2024,GaussianPro,PRoGS}. These approaches selectively transmit only the necessary Gaussians based on rendering views or fluctuating network conditions. However, they do not consider compression. Given the large size of 3DGS, even partial transmission incurs non-negligible bit consumption. 
To address this, GoDe~\cite{gode} integrates compression into progressive 3DGS by employing a learned mask to define Level-of-Detail (LoD) representations and compressing them using codebooks. However, it does not refine Gaussians from shallower levels as they propagate to deeper levels, resulting in a quality mismatch when fine details are required. 
In this work, we propose \textbf{PCGS}, a method that jointly optimizes both the quantity and the quality of anchors. This ensures that anchors decoded at shallower levels are progressively refined to meet the fidelity requirements of deeper levels, enabling efficient and progressive 3DGS compression.

\section{Method}

\subsection{Preliminaries}
\textbf{Scaffold-GS.}
Built on \textbf{3DGS}~\cite{3DGS}, Scaffold-GS~\cite{scaffold} introduces anchors to cluster nearby Gaussians and neural predicts their attributes from anchors. Specifically, each anchor consists of a location $\bm{x}^\text{a} \in\mathbb{R}^3$ and its associated attributions $\left\{{\bm{f}^a\in}\mathbb{R}^{D^\text{a}}, \bm{l}\in\mathbb{R}^6, \bm{o}\in\mathbb{R}^{3\times K}\right\}$, which represent anchor feature, scaling and learnable offsets, respectively. Scaffold-GS employs MLPs along with scaling-based regularization to deduce Gaussian attributes for rendering. Each Gaussian is formulated as:
\begin{equation}
    G(\bm{x}) = \exp{\left(-\frac{1}{2}(\bm{x}-\bm{x}^\text{g})^\top\bm{\Sigma}^{-1}(\bm{x}-\bm{x}^\text{g})\right)}\;,
\end{equation}
where $\bm{x} \in \mathbb{R}^{3}$ is a random 3D location, $\bm{x}^\text{g} \in\mathbb{R}^3$ is the Gaussian location (mean), and $\bm{\Sigma} = \bm{R}\bm{S}\bm{S}^\top\bm{R}^\top$ is the covariance matrix with $\bm{S} \in \mathbb{R}^{3 \times 3}$ and $\bm{R} \in \mathbb{R}^{3 \times 3}$ being scaling and rotation matrices, respectively.
Using rasterization~\cite{ewa}, Gaussians can be splatted to 2D and collectively render the pixel value $\bm{C}\in\mathbb{R}^{3}$ using $\alpha$-composed blending:
\begin{equation}
    \bm{C} = \sum_{i} {\bm{c}_i\alpha_i\prod_{j=1}^{i-1}\left(1-\alpha_j\right)}
\end{equation}
where $\alpha\in\mathbb{R}$ denotes the opacity of each Gaussian after 2D projection and $\bm{c}\in\mathbb{R}^3$ is the view-dependent color~\cite{3DGS}.

\noindent\textbf{HAC++}.
Based on Scaffold-GS, HAC++~\cite{HAC, HAC++} leverages a binary hash grid $\mathcal{H}$ to capture correlations among anchors $\bm{x}^\text{a}$. The hash feature $\bm{f}^\text{h}$ is obtained via interpolation and used to predict both the quantization step and the Gaussian distribution parameters of anchor attributes for context-based entropy modeling. Additionally, HAC++ employs learnable masks to prune ineffective Gaussians and anchors, significantly enhancing the efficiency of 3DGS compression. It achieves SoTA single-rate compression performance. In this work, we improve HAC++ by introducing progressive compression, further expanding its applicability to varying bandwidth and storage constraints.

\subsection{Overview of PCGS}
The framework of PCGS is illustrated in Fig.~\ref{fig:main_method}. It enables the progressive compression of 3DGS by controlling anchors from both quantity and quality perspectives.
\textbf{For quantity}, we introduce a rate-aware progressive masking strategy, as depicted in the lower-left part of Fig.~\ref{fig:main_method}. Specifically, at each progressive level, each anchor learns both an anchor-level mask~$\bm{m}^\text{a}_s$ and a Gaussian-level mask~$\bm{m}^\text{g}_s$, which are derived from the combination of a level-specific masking feature~$\bm{f}^\text{m}_s$ and a base masking feature~$\bm{f}^\text{m}_{base}$. Both masking features are explicitly defined as learnable parameters. At each level $s$, according to the $i$-th anchor mask $\bm{m}[i]^\text{a}_s$, it will either (1) remain undecoded when $\bm{m}[i]^\text{a}_s=0$, (2) be newly decoded when $\bm{m}[i]^\text{a}_s$ transits from $0$ to $1$ at this level, or (3) be refined if it was previously decoded at an earlier level, as illustrated in the right part of Fig.~\ref{fig:main_method}. To ensure a smooth progression, the mask values are carefully designed to be monotonically non-decreasing across progressive levels (detailed in Sec.~\ref{sec:secProMasking}). As the compression levels deepen, more anchors and Gaussians are incorporated, thus improving rendering fidelity.
\textbf{For quality}, we propose a progressive quantization strategy that continuously refines anchor values as levels progress deeper, enhancing fidelity (detailed in Sec.~\ref{sec:ProQ}). As shown in the right part of Fig.~\ref{fig:main_method}, when an anchor is first decoded at a level, its attribute is quantized using \texttt{Round}. At subsequent levels, the previously quantized value is refined through trit-plane quantization~\cite{trit}. For entropy modeling, we use a Gaussian distribution when an anchor is first decoded, as it undergoes \texttt{Round} quantization. For subsequent levels, we adopt a trinomial distribution, which aligns with trit-plane quantization to effectively leverage level-wise context.
By integrating these two strategies, each progressive level improves fidelity by both introducing new anchors and enhancing the precision of existing ones, striking an effective balance for progressive compression.
For simplicity, we omit the anchor index $[i]$ in the following sections. \textit{Unless otherwise specified, all operations are performed per anchor.}

\subsection{Progressive Masking for Quantity Increase}
\label{sec:secProMasking}

Masking strategies have proven effective in 3DGS compression by explicitly reducing the number of parameters (\eg, anchors or Gaussians)~\cite{HAC++, Lee}. By adjusting the masking ratio, a balance between size and fidelity can be achieved. To enable progressive compression, we ensure that the valid ratio of anchors (or Gaussians) is monotonically non-decreasing as the levels progress. This allows newly decoded anchors to be integrated with existing ones, thereby improving rendering fidelity.

However, due to the unstructured nature of anchors, manually defining their importance and consequently determining the order of progressive masking are challenging.
To address this issue, we introduce learnable masks that adaptively determine the progressive level $s$ at which an anchor (or Gaussian) is initially decoded. Specifically, the masks $\bm{m}^\text{g}_s\in\mathbb{R}^{K}$ are first defined at the Gaussian level as
\begin{equation}
    \bm{m}^\text{g}_s= {\mathbbm{1}}[ \texttt{Sig}(\bm{f}_{\text{base}}^\text{m} + \sum_{l=1}^s\texttt{Sfp}(\bm{f}^\text{m}_l)) > \epsilon^{\text{m}}], 
    \label{eq:masking}
\end{equation}
where \texttt{Sig} and \texttt{Sfp} denote the sigmoid and softplus activation functions, respectively, and $\epsilon^{\text{m}}$ is a constant threshold. Here, $\bm{f}_{\text{base}}^\text{m}\in\mathbb{R}^K$ and $\bm{f}^\text{m}_s\in\mathbb{R}^K$ represent the base masking feature and the masking feature at level $s$, which are explicitly defined as learnable parameters. 
The gradient of $\bm{m}^\text{g}_s$ is backpropagated using STE~\cite{Lee,STE}. Notably, in \eqref{eq:masking}, we apply a softplus activation to $\bm{f}^\text{m}_s$ to ensure a positive output. By summing it with outputs from previous levels, we enforce a monotonically non-decreasing mask with respect to levels $s$. This guarantees once a Gaussian is decoded at any level $s$ (\ie, $\bm{m}^\text{g}_s = 1$), it remains valid in all subsequent levels, ensuring reusability.

Building on the Gaussian-level mask, we further derive an anchor-level mask $\bm{m}^\text{a}_s\in\mathbb{R}$ by determining whether at least one valid Gaussian exists within an anchor, following the approach in \cite{HAC++}.
Notably, $\bm{m}^\text{a}_s$ inherits the monotonically non-decreasing property of $\bm{m}^\text{g}_s$.
This monotonicity plays a crucial role in progressive compression. On one hand, it encourages anchors from previous levels to collaborate with newly introduced anchors at the current level. On the other hand, it is fundamental to the progressive quantization strategy (detailed in Subsec.~\ref{sec:ProQ}), as it guarantees that an anchor undergoing refinement can always access its coarse value from the previous level for a finer quantization and context modeling.
By comparing the masks across levels, we can determine whether an anchor is newly decoded or refined from an existing one, which is formulated as:
\vspace{-10pt}
\begin{equation}
\begin{aligned}
    \Delta\bm{m}^{\text{a}}_s &= \bm{m}^{\text{a}}_s - \bm{m}^{\text{a}}_{s-1} \\
\end{aligned}
\label{eq:delta}
\end{equation}
where $\bm{m}^{\text{a}}_{0}$ is set to $0$. 
If $\Delta\bm{m}^{\text{a}}_s=1$, then the corresponding anchor is newly decoded at level $s$; otherwise, it is either previously decoded ($\bm{m}^{\text{a}}_{s-1}=1$) or invalid ($\bm{m}^{\text{a}}_s=0$). Note that $\Delta\bm{m}^{\text{g}}_s$ can be calculated in the same way.
However, merely increasing the number of anchors or Gaussians is insufficient, as the anchors decoded in shallower levels may lack the finer details required for the improved model fidelity at deeper levels. To address this, we further refine the quantization accuracy of existing anchors through progressive quantization.

\subsection{Progressive Quantization and Context for Quality Enhancement}
\label{sec:ProQ}

\begin{figure}[t]
    \centering
    \includegraphics[width=1\linewidth]{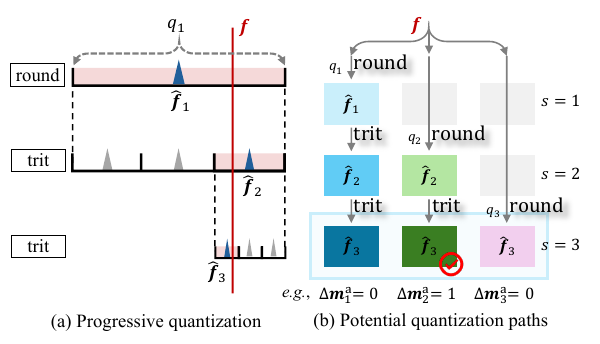}
    \caption{\textbf{Progressive quantization}.
    (a): Illustration of progressive quantization. The real value $\bm{f}$ is initially quantized at level $s=1$ using $\texttt{Round}$ with the step size $q_{1}$. It is then progressively refined using trit-plane quantization at levels $s=2$, and $3$, gradually approaching $\bm{f}$.
    (b): Three possible paths to reach the quantized value $\hat{\bm{f}}_3$ at the current level ($s=3$) in training with $\Delta\bm{m}^\text{a}_2=1$ indicating that the middle path is selected by mask weighting.
    }
    \label{fig:triangle}
    \vspace{-15pt}
\end{figure}

A finer quantization of anchor attributes maintains precision, enhancing fidelity, but also increases storage size due to the complexity of predicting fine-grained values.
In our approach, a value is first decoded at an initial level and then progressively refined at subsequent levels. Upon initial decoding, the value is quantized using \texttt{Round}, with its entropy modeled by a Gaussian distribution. In later levels, the value undergoes refinement by leveraging strong prior information from the previously quantized value. To effectively exploit this prior, we adopt trit-plane quantization and model entropy using a trinomial distribution, as shown in Fig.~\ref{fig:triangle} (a).

\noindent\textbf{For an anchor at its initial decoding}, there is no prior quantization information available. Thus, the value $\bm{f}$ is quantized using $\texttt{Round}$, \ie $\hat{\bm{f}}_s = \texttt{Round}(\bm{f}, q_s)$, where the quantization step $q_s$ is determined by an MLP from the hash feature $\bm{f^\text{h}}$,
\vspace{-10pt}
\begin{equation}
    q_s, \bm{\mu}_s, \bm{\sigma}_s=\text{MLP}\left(\bm{f}^\text{h}, \lambda_s\right)
\label{eq:round_step}
\end{equation}
where $\bm{f}^h$ is the hash feature obtained by querying the anchor location $\bm{x}^\text{a}$ in the hash grid $\mathcal{H}$. The parameter $\lambda_s$ controls the R-D loss trade-off (detailed in \eqref{eq:loss}). $\bm{\mu}_s$ and $\bm{\sigma}_s$ represent the estimated Gaussian distribution parameters for entropy modeling.
Since it is at its initial decoding, the quantized values lack prior information and can be estimated as any value within the real number space. Thus, we model the probability of the quantized result using a Gaussian distribution $\phi$, which imposes no preconditions on the value range:
\begin{equation}
\begin{aligned}
    p^\text{G}(\hat{\bm{f}}_s)
    &= \int_{\hat{\bm{f}}_s-\frac{\bm{q}_s}{2}}^{\hat{\bm{f}}_s+\frac{\bm{q}_s}{2}}\phi\left(x\mid {\bm{\mu}_s, \bm{\sigma}_s}\right)\,dx, 
\end{aligned}
\label{eq:gaussian_prob}
\end{equation}
where $\hat{\bm{f}}_s$ is obtained by adding noise to $\bm{f}$ during training to preserve gradient flow, and $\texttt{Round}$ during testing~\cite{HAC++}.

\noindent\textbf{For an anchor that has been previously decoded}, it is refined at the current level. Given the previously decoded value $\hat{\bm{f}}_{s-1}$ and its quantization step size $q_{s-1}$, the real value $\bm{f}$ is known to lie within range of $[\hat{\bm{f}}_{s-1}-\frac{q_{s-1}}{2}, \hat{\bm{f}}_{s-1}+\frac{q_{s-1}}{2})$. To fully utilize this prior, we adopt trit-plane quantization~\cite{trit}, as illustrated in Fig.~\ref{fig:triangle} (a). Specifically, trit-plane quantization refines the value by subdividing the range into three equal sub-intervals with midpoints: $\{\hat{\bm{f}}_{s^c}\}_{c=1,2,3}=\{\hat{\bm{f}}_{s-1}-\frac{q_{s-1}}{3}, \hat{\bm{f}}_{s-1}, \hat{\bm{f}}_{s-1}+\frac{q_{s-1}}{3}\}$ (\ie, the three triangles in the second row in Fig.~\ref{fig:triangle} (a)). The quantized value $\hat{\bm{f}}_{s}$ is assigned to the midpoint closest to $\bm{f}$:
\begin{equation}
    \hat{\bm{f}}_{s} =  \hat{\bm{f}}_{s^u}, \quad \text{where} \quad u=\text{arg}\min_c|\bm{f}-\hat{\bm{f}}_{s^c}|.
\label{eq:trit_plane}
\end{equation}
If $\bm{f}$ lies exactly at a boundary between two sub-intervals, the value on the right is chosen. The quantization step is then updated as $q_{s}=\frac{q_{s-1}}{3}$ for recursive use of tirt-plane quantization in subsequent levels.

For entropy modeling, we employ a trinomial distribution, which aligns naturally with the three candidate values in trit-plane quantization. Given the strong prior from level $s-1$, we predict the probability via:
\begin{equation}
\begin{aligned}
    p^\text{T}({\hat{\bm{f}}_s})&=\bm{p}_{s^u},\\
    \{\bm{p}_{s^c}\}_{c=1,2,3}&=\text{MLP}(\hat{\bm{f}}_{s-1}, \bm{f}^h,\lambda_s).
\end{aligned}
\label{trinomial_prob}
\end{equation}
Note that although anchor values in subsequent levels could also be quantized using $\texttt{Round}$ and their entropy modeled by a Gaussian distribution, this approach does not fully exploit the prior information from previous levels, leading to suboptimal performance. Please refer to ablation study for evaluations.

\noindent\textbf{An anchor may be initially decoded at different levels}.
Trit-plane quantization is a causal process that relies on the anchor’s value at its first decoded level. However, due to the progressive masking strategy, during training an anchor may be initially decoded at different levels with varying quantization steps, potentially leading to different trit-plane quantized results. This is illustrated in Fig.~\ref{fig:triangle} (b) as a lower triangular matrix, where different columns represent the anchor is first decoded at a different level (\ie, $s=1, 2$, or $3$) and it leads to three different $\hat{\bm{f}}_3$ in training. 

This is not a problem during testing, where the anchor's initial decoding level can be directly obtained using the trained $\Delta\bm{m}^\text{a}$, and hence the desired $\hat{\bm{f}}_s$ can be decoded progressively. However, during training, since $\Delta\bm{m}^\text{a}$
is still learnable, all possible $\hat{\bm{f}}_s$ must be enumerated and weighted by the mask $\Delta\bm{m}^\text{a}$ to enable the joint optimization of both the mask and the quantized value.
Here, enumerating all possible quantization values $\hat{\bm{f}}_s$ is complex and increases the learning difficulty. To address this issue, for the quantization step when an anchor is firstly decoded (\ie, the \texttt{Round} steps at the diagonal of Fig.~\ref{fig:triangle}~(b)), we define it as $q_{s} = \frac{q_{1}}{3^{s-1}}$. Given the candidate values in the trit-plane as $\{\hat{\bm{f}}_{{s}^c}\}_{c=1,2,3} = \{\hat{\bm{f}}_{s-1} - \frac{q_{s-1}}{3}, \hat{\bm{f}}_{s-1}, \hat{\bm{f}}_{s-1} + \frac{q_{s-1}}{3}\}$, this formulation guarantees that the quantized value at the current level $s$ remains identical, regardless of the level at which the anchor is initially decoded.
This approach unifies the different paths illustrated in Fig.~\ref{fig:triangle} (b) into a single path: during training, we directly apply $\texttt{Round}$ to $\bm{f}$ using $q_{s}$ at level $s$ to obtain $\hat{\bm{f}}_s$, making it independent of $\Delta\bm{m}^\text{a}_s$.

\subsection{Training Losses for Progressive Compression}
\label{sec:Collaboration}
As the level progresses deeper, new anchors and Gaussians are decoded, while existing anchors are refined, collectively enhancing the model’s fidelity progressively. Specifically, at each training iteration, we randomly sample a level $s$, and compute the entropy loss based on the \textit{incremental} bit consumption at this level.
For the first level ($s=1$), the loss formulation follows HAC++~\cite{HAC++}. For levels $s \geq 2$, the entropy constraint $L_{\text{entropy}}$ is applied from an \textit{incremental} perspective:
\vspace{-5pt}
\begin{small}
\begin{equation}
    \begin{aligned}
        L_{\text{entropy}} = \;
        & \sum_i^N b[i]_{s},~~~\; \text{where for each anchor $i$}, \\
        b_{s} = \;
        & \bm{m}^\text{a}_{s}\Delta\bm{m}^\text{a}_{s}\sum_{\bm{f}\in\{\bm{f}^a, \bm{l}\}} \sum_{j=1}^{D^f} \left(-\log_2 p^\text{G}(\hat{{f}}_{s,j})\right) + \\
        & \bm{m}^\text{a}_{s}(1-\Delta\bm{m}^\text{a}_{s})\sum_{\bm{f}\in\{\bm{f}^a, \bm{l}\}} \sum_{j=1}^{D^f} \left(-\log_2 p^\text{T}(\hat{{f}}_{s,j})\right) + \\
        & \sum_{k=1}^{K}\Delta\bm{m}^\text{g}_{s,k}\sum_{j=1}^{3} \left(-\log_2 p^\text{G}(\hat{{o}}_{s,3k+j})
        \right)
    \end{aligned}
\end{equation}
\end{small}
\noindent where the first and second terms respectively account for the bits needed for a newly decoded anchor at level-$s$ (with $\Delta\bm{m}^\text{a}_s=1$), modeled as Gaussian distribution in \eqref{eq:gaussian_prob}, and the bits needed for a refined anchor (with $\Delta\bm{m}^\text{a}_s=0$), modeled as trinomial distribution in \eqref{trinomial_prob}, and the third term accounts for the bits needed for any newly added Gaussians (with $\Delta\bm{m}^\text{g}_{s,k}=1$) associated with the anchor.
Note that for Gaussian offsets, only the quantity increases while previously decoded values remain unchanged.

The final loss at level $s$ becomes
\begin{equation}
    Loss = L_{\text{Scaffold}} + \lambda_s\frac{1}{N(D^a+6+3K)}(L_{\text{entropy}} + L_{\text{hash}}).
    \label{eq:loss}
\end{equation}
where $L_{\text{Scaffold}}$ and $L_{\text{hash}}$ are the same as those defined in HAC++~\cite{HAC++} and the trade-off parameter $\lambda_s$ varies with $s$ to balance the R-D performance at different levels.

\begin{figure*}[t]
    \centering
    \includegraphics[width=1.0\linewidth]{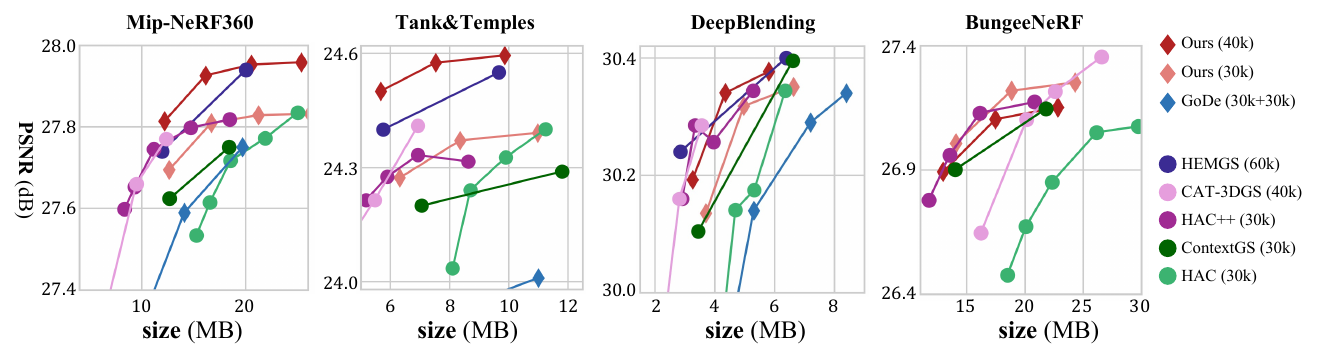}%
    \vspace{-15pt}
    \caption{\textbf{R-D curve comparison of different methods.} The number of training iterations for each method is indicated in parentheses. For GoDe, it applies an additional finetune stage of $30k$ iterations to the Scaffold-GS which is originally trained for $30k$. Diamond markers \scalebox{1.2}{$\Diamond$} represent \textbf{progressive compression} methods, while circle markers $\bigcirc$ denote traditional \textbf{single-rate compression} methods. More results can be obtained in the Appendix.}
    \vspace{-10pt}
    \label{fig:main_results}
\end{figure*}

\section{Experiments}

\subsection{Implementation Details}
PCGS is implemented using the PyTorch~\cite{pytorch} framework, building upon the HAC++~\cite{HAC++} repository. The hyperparameters remain consistent with HAC++, except that we employ a set of $\lambda_s$ to train the model progressively within a single training process. At each iteration, we randomly sample a level $s$ and its corresponding $\lambda_s$ from this set for training. Since all rates are covered within one training session, we appropriately extend the training iterations to $40k$ to ensure sufficient optimization. Additionally, we also present the results trained by $30k$ iterations for a fair comparison with the standard protocol.

\subsection{Experiment Evaluation}
\textbf{Baselines}. Progressive compression of 3DGS still lacks exploration. Specifically, GoDe~\cite{gode} organizes Gaussians into layers and applies codebooks for progressive compression. 
Other compression methods generate only a single rate per training. By adjusting training parameters and retraining the scene multiple times, we obtain the R-D curves for HAC~\cite{HAC}, HAC++\cite{HAC++}, ContextGS\cite{Contextgs}, HEMGS~\cite{HEMGS}, and CAT-3DGS~\cite{CAT3DGS}. We present the curves of these approaches in Fig.~\ref{fig:main_results}, as they exhibit the best compression performance among existing methods.
For all metrics (\ie, PSNR, SSIM~\cite{ssim}, LPIPS~\cite{LPIPS}, size, training time, and encoding/decoding time) and results of more comparison methods (including the baselines 3DGS~\cite{3DGS} and Scaffold-GS~\cite{scaffold}, and various compression methods~\cite{RDOGaussian, SOG, Compact3d, Lightgaussian, EAGLES, liu2024compgs, Lee, Simon, papantonakis, MesonGS}), please refer to the Appendix.

\noindent\textbf{Dataset}. We evaluate our PCGS on the large-scale real-world scenes including Mip-NeRF360~\cite{mip360}, DeepBlending~\cite{deepblending}, Tanks\&Temples~\cite{tant}, and BungeeNeRF~\cite{BungeeNeRF} datasets, which can better highlight the advantages of progressivity due to their large volumes.

\noindent\textbf{Results}.
As shown in Fig.~\ref{fig:main_results}, our PCGS, despite being trained only once to obtain the progressive R-D curve, achieves compression performance comparable to SoTA single-rate methods.
More importantly, PCGS enables compression into different progressive bitstreams, efficiently adapting to dynamic network and storage constraints.
Compared to the progressive method GoDe, which undergoes $60k$ training iterations in total to achieve progressivity, our PCGS still outperforms it. This is because GoDe solely increases the number of anchors for progressivity, whereas PCGS jointly optimizes both the quantity and quality of anchors, leading to superior progressive performance.
Additionally, PCGS effectively exploits contextual correlations across levels to model entropy more accurately, further compacting the incremental bitstreams. 
Notably, additional training iterations do not improve performance on the BungeeNeRF dataset due to its scale discrepancy between training and testing views, which lead to overfitting.
When compared to single-rate approaches, PCGS surpasses HEMGS on most datasets, even with fewer training iterations. Note that PCGS only needs $30/40k$ iterations to obtain the entire R-D curve while HEMGS needs $60k$ iterations for \emph{each} sampled rate, which further demonstrates the effectiveness and efficiency of our context designs.

\noindent\textbf{Decoding Process}. PCGS enables a progressive decoding process by reusing existing bitstreams. Initially, shared components such as MLPs, anchor locations $\bm{x}^\text{a}$, the hash grid $\mathcal{H}$, and masks $\bm{m}^\text{g}$ (from which $\bm{m}^\text{a}$ is derived) are decoded from the header information. 
Note that since anchor locations are encoded using GPCC~\cite{GPCC}, which relies on mutual information among anchors to reduce entropy, we encode all valid anchor locations together in the header, rather than encoding the incremental anchor locations at each level, to reduce the total bit consumption.
Based on the mask information, the model determines which anchors to be newly decoded or refined at each level.
Specifically, at the first level ($s=1$), attributes of valid anchors and Gaussian offsets are decoded using Gaussian distributions. At the next level ($s=2$), two operations occur: (1) already decoded anchors are refined using the progressive quantization mechanism, with a trinomial distribution for probability estimation, and (2) newly introduced anchors and Gaussian offsets are decoded via a Gaussian distribution. This process continues consistently across all progressive levels, ensuring efficient and structured decoding.

\begin{figure} [t]
    \centering   \includegraphics[width=1.00\linewidth]{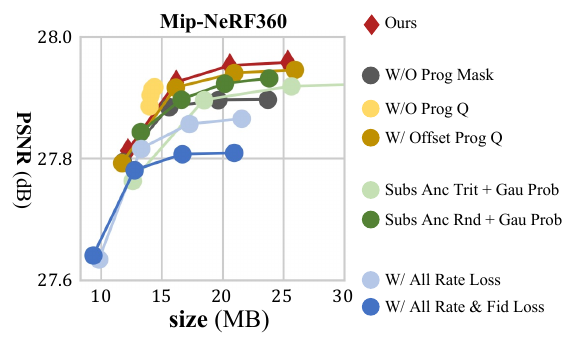}
    \vspace{-20pt}
    \caption{\textbf{R-D curve results of ablation study.} Experiments are conducted on Mip-NeRF360 \cite{mip360}.}
    \label{fig:ablation} 
\vspace{-15pt}
\end{figure}

\subsection{Ablation Study}
\label{sec:ablation}

We conduct ablation studies on the Mip-NeRF360 dataset~\cite{mip360} with sufficient training iterations of $40k$, as it deals with the most diverse set of scenes and yields the most convincing results. The ablations are analyzed from three perspectives: (1) the effectiveness of progressivity, (2) approaches to quantize and entropy model anchors decoded in subsequent levels, and (3) the design of the training loss.
Results are shown in Fig.~\ref{fig:ablation}.

\noindent\textbf{Effectiveness of Progressivity.} Disabling progressive masking (W/O Prog Mask, \ie, keeping the mask identical across levels) prevents the model from improving fidelity in the high-rate segments due to the lack of additional anchors or Gaussians. Additionally, when the progressive quantization mechanism is removed (W/O Prog Q), progressivity relies solely on the mask, resulting in a narrow rate adjustment range with limited fidelity improvements, as the quality of existing anchors remains fixed and fails to adapt to varying levels of details. Furthermore, applying progressive quantization to offsets (W Offset Prog Q) shifts the model’s focus toward optimizing offset sizes, which may increase overall rate consumption and impact fidelity.

\noindent\textbf{Quantization and Entropy Modeling of Subsequent Anchors.} For anchors that have been previously decoded (\ie, not at their initial decoding), replacing the trinomial distribution with a Gaussian distribution (Subs Anc Trit + Gau Prob) leads to inaccurate probability estimation, as it fails to sufficiently exploit the quantization context across levels. Similarly, if progressive quantization of these subsequent-level anchors is replaced with direct $\text{Round}$ and entropy modeling using a Gaussian distribution (Subs Anc Rnd + Gau Prob), the performance degrades due to ineffective adaptation to progressive refinement.

\noindent\textbf{Training Loss Design.} Using the accumulated entropy from all previous and current levels as rate regularization (W/ All Rate Loss) causes the model to degrade in fidelity, as it does not explicitly consider the quality of reconstructions at earlier levels when optimizing the current level. However, if fidelity at previous levels is incorporated into the optimization (W/ All Rate \& Fid Loss), the model’s performance becomes even worse, as this introduces excessive complexity and hinders the model from effectively balancing the R-D trade-off across levels.

\begin{figure} [t]
    \centering   \includegraphics[width=1.0\linewidth]{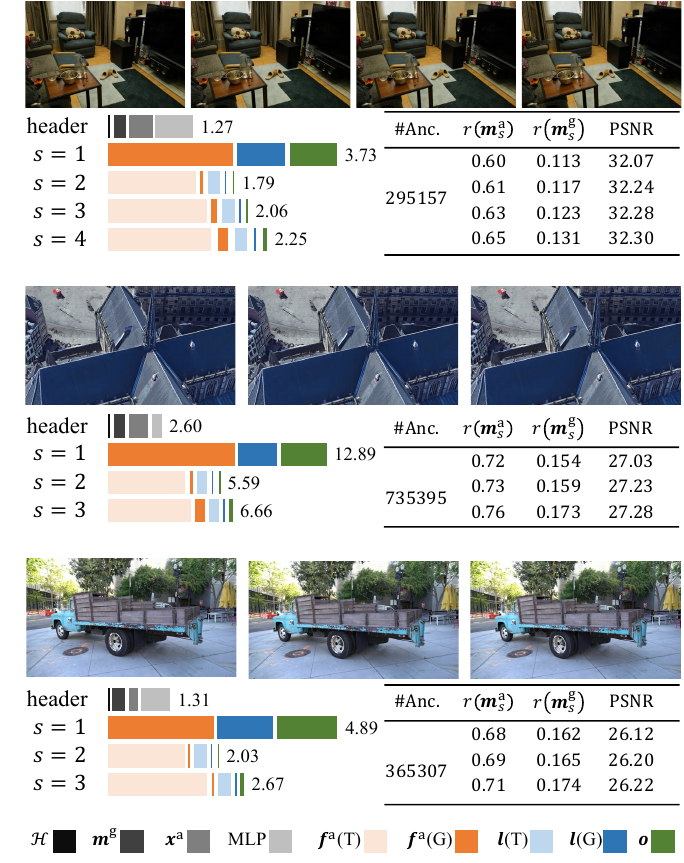}
    \vspace{-15pt}
    \caption{\textbf{Relative size relations of different components.} Total size of each header or level is at the right of each bin, measured in MB.
    \textbf{Upper part} of each chunk: Qualitative results of the \textit{room}, \textit{amsterdam}, and \textit{truck} scenes across different levels. \textbf{Lower part} of each chunk: Statistical data showing the size and mask information across different levels. $r(\bm{m}^\text{a}_s)$ and $r(\bm{m}^\text{g}_s)$ represent the valid mask ratios for anchors and Gaussians, respectively. The color notations are explained in the \textbf{bottom} of the figure. (T) and (G) denote the attribute is refined or newly decoded, respectively.}
    \label{fig:progressive_bit_allocation} 
\vspace{-10pt}
\end{figure}

\subsection{Progressive Bit Allocations}
To achieve progressive compression, we encode and decode the scene level by level. The statistical information on bit allocations and mask ratios across levels is shown in Fig.~\ref{fig:progressive_bit_allocation}. Specifically, we first encode and store the global header information, which includes MLPs, anchor locations, the hash grid, and masks. 
Additionally, we present the relative bit consumption at each level.
For anchors at levels $s\geq2$, the bit consumption consists of two parts: bits from trit-plane refinement and bits from newly decoded anchors and Gaussian offsets. While the bits from newly decoded anchors and Gaussian offsets share a small portion, they play a crucial role in improving fidelity and enhancing the progressivity of the model.
Moreover, both $r(\bm{m}^\text{a}_s)$ and $r(\bm{m}^\text{g}_s)$ increase as the level deepens, indicating more new anchors and Gaussians are decoded, which highlights the effectiveness of the progressive masking strategy.

\begin{figure} [t]
    \centering   \includegraphics[width=0.9\linewidth]{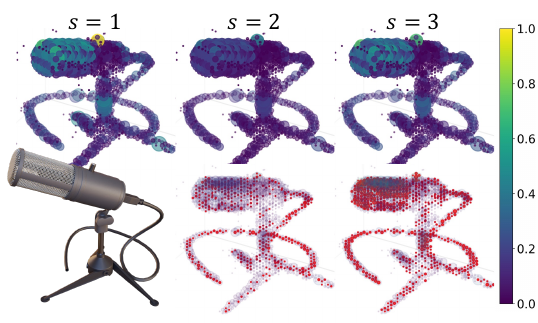}
    \vspace{-10pt}
    \caption{\textbf{Visualization of bit allocation across progressive levels in the \textit{mic} scene}~\cite{NeRF}. Following HAC++~\cite{HAC++}, we voxelize the scene, with each voxel represented by a ball. \textbf{Upper}: The size and color of a ball indicate the number of anchors and the amount of \textit{incremental} bits (normalized) within it. \textbf{Lower-right}: Red balls indicate the newly decoded anchors at the current level.}
    \label{fig:bit_allocation} 
\vspace{-15pt}
\end{figure}

\subsection{Visualization}
Following HAC++~\cite{HAC++}, we visualize the bit allocation of PCGS in the \textit{mic} scene, as shown in Fig.\ref{fig:bit_allocation} (\textbf{upper}). At the first level (\ie, $s=1$), the highest number of bits is consumed, as it provides basic information for subsequent levels. Additionally, the Gaussian distribution used at this level lacks prior information, making probability estimation more challenging. At higher levels (\ie, $s=2$ and $3$), they consume fewer bits, which demonstrates the effectiveness of the trinomial distribution. 
Moreover, as shown in Fig.~\ref{fig:bit_allocation} (\textbf{lower-right}), new anchors are decoded at subsequent levels. Complex areas with higher bit consumptions activate more new anchors for further refinement of fidelity.

\section{Conclusion}
In this paper, we have presented PCGS, a novel framework for progressive compression of 3DGS. By controlling \textbf{both the quantity and quality} of anchors, PCGS achieves performance comparable to SoTA single-rate compression methods. Extensive experiments have demonstrated the effectiveness of the proposed components. More importantly, the progressive nature of PCGS makes it well-suited for on-demand applications, where dynamic bandwidth and diversion storage conditions often arise, significantly broadening the applicability of 3DGS.

\newpage
{
    \small
    \bibliographystyle{ieeenat_fullname}
    \bibliography{main}
}

\clearpage

\setcounter{page}{1}
\renewcommand{\thetable}{\Roman{table}}
\renewcommand{\thesection}{\Alph{section}}
\onecolumn
\setcounter{section}{0}
\setcounter{table}{0}
\setcounter{figure}{0}

\begin{center}
    {\LARGE \textbf{PCGS: Progressive Compression of 3D Gaussian Splatting}} \\[3mm]
    {\large Supplementary Material} \\[8mm]
\end{center}

\section{More Quantitative Results of PCGS}
\label{sec:result_each_scene}

In this section, we firstly present our quantitative results (trained for $40k$ iterations) for each scene across various datasets, including Mip-NeRF360~\cite{mip360}, DeepBlending~\cite{deepblending}, Tanks\&Temples~\cite{tant}, and BungeeNeRF~\cite{BungeeNeRF}, as shown in Tables~\ref{tab:tandt}, \ref{tab:mip360}, \ref{tab:DeepBlending}, and \ref{tab:BungeeNeRF}, respectively. Furthermore, we present results on the small-scale Synthetic-NeRF dataset~\cite{NeRF}. While this dataset is less suitable for progressive compression due to its limited size, we include it to ensure the completeness of our evaluations. Notably, the overhead introduced by MLPs (approximately $0.5$ MB for this dataset) becomes more pronounced in this dataset. Despite this challenge, our method still achieves relatively strong compression performance.

For the first level (\ie, $s=1$) of each scene, in addition to entropy coding of anchor attributes (\ie, $\bm{f}^\text{a}$, $\bm{l}$, and $\bm{o}$), the encoding process also includes storing and coding of header information, \ie, MLPs, anchor locations $\bm{x}^\text{a}$, the hash grid $\mathcal{H}$, and Gaussian-level masks $\bm{m}^\text{g}$ (from which the anchor-level masks $\bm{m}^\text{a}$ can be derived). This results in longer encoding and decoding times and larger sizes at the first level.

For subsequent levels (\ie, $s\geq 2$), the size is calculated as the sum of the previous level’s size and the incremental size (\ie, $\Delta$ Size in the tables) of the current level. The encoding and decoding times are also calculated in an incremental perspective. Benefiting from this progressive design, on-demand applications require only the incremental bits and encoding/decoding times for levels $s \geq 2$, underscoring the advantages of the progressive compression pipeline.

Additionally, we also provide results of PCGS trained by $30k$ iterations to align with the standard training protocol, as shown in Table~\ref{tab:30k_results}. Notably, even with only $30k$ iterations, PCGS has still achieved excellent compression performance.

\section{Quantitative Results of Other Methods}
\label{sec:more_results}
In this section, we provide more quantitative results of comparison methods, including baselines 3DGS~\cite{3DGS} and Scaffold-GS~\cite{scaffold}, as well as various compression methods~\cite{HAC, HAC++, Contextgs, HEMGS, CAT3DGS, SOG, Compact3d, Lightgaussian, liu2024compgs, Lee, Simon, papantonakis, RDOGaussian, MesonGS, EAGLES}. As shown in Tables~\ref{tab:main_quantitative}, our PCGS, despite being
trained only once to obtain the progressive R-D curve,
achieves compression performance comparable to SoTA single-rate methods.
\vspace{30pt}

\begin{table*}[ht]
\begin{small}
\centering
    \begin{tabular}{c|c|ccccc|c|cc}
    \toprule[2pt]
    Scenes  & $\lambda_s$ & PSNR$\uparrow$ & SSIM$\uparrow$ & LPSPI$\downarrow$ &$\Delta$ Size$\downarrow$ &Size$\downarrow$ &Train time &Enc  time &Dec time \\ \toprule
    \multirow{3}{*}{Train} 
    &$8e-4$ &22.88 & 0.8191 & 0.2195 & 5.14 & 5.14  & \multirow{3}{*}{2386} & 5.5  & 8.2   \\
    &$4e-4$ &22.95 & 0.8230 & 0.2155 &1.68 & 6.82  &  & 1.1  & 1.2   \\
    &$0.5e-4$ &22.97 & 0.8239 & 0.2143 &2.02 & 8.84  &   & 1.2  & 1.4   \\ \hline
    
    \multirow{3}{*}{Truck}
    &$8e-4$ &26.12 & 0.8850 & 0.1520 & 6.20 & 6.20  & \multirow{3}{*}{2508} & 7.3  & 10.9   \\
    &$4e-4$ &26.20 & 0.8877 & 0.1491 &2.03 & 8.23  &   & 1.5  & 1.6   \\
    &$0.5e-4$ &26.22 & 0.8883 & 0.1482 &2.67 & 10.90  &  & 1.7  & 2.0   \\ \hline
    \multirow{3}{*}{\textbf{Avg.}} 
    &$8e-4$ &24.50 & 0.8521 & 0.1857 & 5.67  & 5.67  & \multirow{3}{*}{2447} & 6.4  & 9.6   \\
    &$4e-4$&24.58 & 0.8554 & 0.1823 & 1.85  & 7.52  &  & 1.3  & 1.4   \\
    &$0.5e-4$&24.59 & 0.8561 & 0.1813 & 2.34  & 9.87  &  & 1.5  & 1.7   \\
    \toprule[2pt]
    \end{tabular}
    \vspace{-8pt}
    \caption{Our results of each scene on the \textbf{Tanks\&Temples} dataset~\cite{tant} trained with $40k$ iterations. Times are measured in Second (s) and sizes are measured in MegaByte (MB). The $\Delta$ Size represents the incremental size of each level. }
    \label{tab:tandt}
\end{small}
\end{table*}

\begin{table*}[ht]
\small
\centering
    \vspace{-6pt}
    \begin{tabular}{c|c|ccccc|c|cc}
    \toprule[2pt]
    Scenes  & $\lambda_s$ & PSNR$\uparrow$ & SSIM$\uparrow$ & LPSPI$\downarrow$ &$\Delta$ Size$\downarrow$ &Size$\downarrow$  &Train time &Enc time &Dec time \\ \toprule
     \multirow{3}{*}{Drjohnson} 
    &$8e-4$ &29.70 & 0.9045 & 0.2620 & 3.73 & 3.73  & \multirow{3}{*}{2218} & 4.0  & 5.6   \\
    &$4e-4$ &29.83 & 0.9069 & 0.2584 &1.29 & 5.02  &   & 0.9  & 1.0   \\
    &$0.5e-4$ &29.85 & 0.9074 & 0.2576 &1.68 & 6.70  &  & 1.0  & 1.2   \\ \hline
    \multirow{3}{*}{Playroom}
    &$8e-4$ &30.69 & 0.9091 & 0.2657 & 2.80 & 2.80  & \multirow{3}{*}{2209} & 2.9  & 4.2   \\
    &$4e-4$ &30.85 & 0.9113 & 0.2620 &0.88  & 3.68  &  & 0.6  & 0.7   \\
    &$0.5e-4$ &30.91 & 0.9119 & 0.2609 &1.24 & 4.92  &  & 0.8  & 1.0   \\  \hline
    \multirow{3}{*}{\textbf{Avg.}}
    &$8e-4$&30.19 & 0.9068 & 0.2639 & 3.26  & 3.26  & \multirow{3}{*}{2214} & 3.5  & 4.9   \\
    &$4e-4$&30.34 & 0.9091 & 0.2602 & 1.09  & 4.35  &  & 0.7  & 0.8   \\
    &$0.5e-4$&30.38 & 0.9097 & 0.2593 & 1.45  & 5.81  &  & 0.9  & 1.1   \\

    \toprule[2pt]
    \end{tabular}
    \vspace{-8pt}
    \caption{
    Our results of each scene on the \textbf{DeepBlending} dataset~\cite{deepblending} trained with $40k$ iterations. Times are measured in Second (s) and sizes are measured in MegaByte (MB). The $\Delta$ Size represents the incremental size of each level.
    }
    \label{tab:DeepBlending}
\end{table*}

\begin{table*}[ht]
\small
\centering
    \vspace{-6pt}
    \begin{tabular}{c|c|ccccc|c|cc}
    \toprule[2pt]
    Scenes  & $\lambda_s$ & PSNR$\uparrow$ & SSIM$\uparrow$ & LPSPI$\downarrow$ &$\Delta$ Size$\downarrow$ &Size$\downarrow$ &Train time &Enc  time &Dec time \\ \toprule

    \multirow{3}{*}{Amsterdam} 
    &$8e-4$ &27.03 & 0.8793 & 0.2028 & 15.49  & 15.49  & \multirow{3}{*}{4862} & 16.3  & 23.9   \\
    &$4e-4$ &27.23 & 0.8861 & 0.1942 & 5.59  & 21.08  &  & 3.3  & 3.7   \\
    &$0.5e-4$ &27.28 & 0.8888 & 0.1891 & 6.66  & 27.74  &  & 4.0  & 5.0   \\  \hline
    \multirow{3}{*}{Bilbo}
    &$8e-4$ &27.91 & 0.8818 & 0.1988 & 12.18  & 12.18  & \multirow{3}{*}{3684} & 12.5  & 17.9   \\
    &$4e-4$ &28.09 & 0.8872 & 0.1903 & 4.35  & 16.53  &  & 2.5  & 2.9   \\
    &$0.5e-4$ &28.11 & 0.8891 & 0.1856 & 5.24  & 21.77  &  & 3.1  & 4.0   \\  \hline
    \multirow{3}{*}{Hollywood}
    &$8e-4$ &24.43 & 0.7657 & 0.3319 & 12.35  & 12.35  & \multirow{3}{*}{3208} & 12.3  & 16.9   \\
    &$4e-4$ &24.58 & 0.7736 & 0.3254 & 3.98  & 16.33  &  & 2.3  & 2.5   \\
    &$0.5e-4$ &24.64 & 0.7774 & 0.3210 & 4.62  & 20.95  &  & 2.7  & 3.3   \\  \hline
    \multirow{3}{*}{Pompidou}
    &$8e-4$ &25.63 & 0.8517 & 0.2347 & 13.87  & 13.87  & \multirow{3}{*}{4510}  & 14.4  & 20.6   \\
    &$4e-4$ &25.81 & 0.8570 & 0.2293 & 4.85  & 18.72  &  & 2.9  & 3.2   \\
    &$0.5e-4$ &25.85 & 0.8585 & 0.2270 & 5.84  & 24.56  &  & 3.4  & 4.3   \\  \hline
    \multirow{3}{*}{Quebec}
    &$8e-4$ &30.13 & 0.9338 & 0.1610 & 10.94  & 10.94  & \multirow{3}{*}{3511}  & 11.2  & 16.1   \\
    &$4e-4$ &30.43 & 0.9380 & 0.1562 & 3.78  & 14.72  &  & 2.2  & 2.5   \\
    &$0.5e-4$ &30.49 & 0.9388 & 0.1546 & 4.47  & 19.18  &  & 2.6  & 3.2   \\  \hline
    \multirow{3}{*}{Rome}
    &$8e-4$ &26.22 & 0.8694 & 0.2110 & 13.01  & 13.01  & \multirow{3}{*}{3902} & 13.2  & 18.8   \\
    &$4e-4$ &26.47 & 0.8757 & 0.2049 & 4.42  & 17.42  &  & 2.6  & 2.9   \\
    &$0.5e-4$ &26.54 & 0.8778 & 0.2018 & 5.30  & 22.72  &  & 3.1  & 3.9   \\  \hline
    \multirow{3}{*}{\textbf{Avg.}}
    &$8e-4$&26.89 & 0.8636 & 0.2234 & 12.97  & 12.97  & \multirow{3}{*}{3946} & 13.3  & 19.0   \\
    &$4e-4$&27.10 & 0.8696 & 0.2167 & 4.50  & 17.47  &  & 2.6  & 2.9   \\
    &$0.5e-4$&27.15 & 0.8717 & 0.2132 & 5.35  & 22.82  &  & 3.2  & 3.9   \\
    
    \toprule[2pt]
    \end{tabular}
    \vspace{-8pt}
    \caption{
    Our results of each scene on the \textbf{BungeeNeRF} dataset~\cite{BungeeNeRF} trained with $40k$ iterations. Times are measured in Second (s) and sizes are measured in MegaByte (MB). The $\Delta$ Size is the incremental size of each level.
    }
    \label{tab:BungeeNeRF}
\end{table*}

\begin{table*}[ht]
\small
\centering
    
    \vspace{-6pt}
    \begin{tabular}{c|c|ccccc|c|cc}
    \toprule[2pt]
    Scenes  & $\lambda_s$ & PSNR$\uparrow$ & SSIM$\uparrow$ & LPSPI$\downarrow$ &$\Delta$ Size$\downarrow$ &Size$\downarrow$ &Train time &Enc  time &Dec time \\ \toprule
    \multirow{4}{*}{Bicycle} 
    &$4e-4$ &25.12 & 0.7387 & 0.2724 & 19.09  & 19.09  & \multirow{4}{*}{5570} & 20.9  & 33.9   \\
    &$2.5e-4$ &25.17 & 0.7409 & 0.2705 & 6.12  & 25.21  &  & 3.8  & 4.0   \\
    &$1e-4$ &25.17 & 0.7415 & 0.2696 & 6.93  & 32.14  &  & 4.0  & 4.5   \\
    &$0.2e-4$ &25.18 & 0.7419 & 0.2688 & 7.59  & 39.73  &  & 4.5  & 5.4   \\  \hline

    \multirow{4}{*}{Garden} 
    &$4e-4$ &27.37 & 0.8416 & 0.1583 & 18.27  & 18.27  & \multirow{4}{*}{5647} & 18.8  & 28.5   \\
    &$2.5e-4$ &27.49 & 0.8452 & 0.1543 & 5.80  & 24.07  &  & 3.4  & 3.6   \\
    &$1e-4$ &27.53 & 0.8463 & 0.1527 & 6.36  & 30.42  &  & 3.6  & 4.1   \\
    &$0.2e-4$ &27.53 & 0.8467 & 0.1517 & 6.82  & 37.24  &  & 4.0  & 4.6   \\  \hline

    \multirow{4}{*}{Stump} 
    &$4e-4$ &26.61 & 0.7605 & 0.2734 & 13.49 & 13.49  & \multirow{4}{*}{3617} & 13.4  & 21.4   \\
    &$2.5e-4$ &26.66 & 0.7622 & 0.2717 & 3.83  & 17.32  &  & 2.3  & 2.4   \\
    &$1e-4$ &26.67 & 0.7626 & 0.2711 & 4.23  & 21.55  &  & 2.4  & 2.7   \\
    &$0.2e-4$ &26.67 & 0.7627 & 0.2707 & 4.64  & 26.18  &  & 2.8  & 3.4   \\     \hline

    \multirow{4}{*}{Room} 
    &$4e-4$ &32.07 & 0.9232 & 0.2094 & 5.00  & 5.00  & \multirow{4}{*}{2914}& 5.3  & 7.9   \\
    &$2.5e-4$ &32.24 & 0.9262 & 0.2043 & 1.79  & 6.79  &  & 1.1  & 1.2   \\
    &$1e-4$ &32.28 & 0.9271 & 0.2021 & 2.06  & 8.85  &  & 1.2  & 1.4   \\
    &$0.2e-4$ &32.30 & 0.9274 & 0.2013 & 2.25  & 11.10  &  & 1.3  & 1.6   \\ \hline
    
    \multirow{4}{*}{Counter} 
    &$4e-4$ &29.75 & 0.9148 & 0.1913 & 6.80  & 6.80  & \multirow{4}{*}{3111} & 6.6  & 9.9   \\
    &$2.5e-4$ &29.88 & 0.9176 & 0.1880 & 2.19  & 8.99  &  & 1.2  & 1.4   \\
    &$1e-4$ &29.91 & 0.9182 & 0.1870 & 2.39 & 11.38  &  & 1.3  & 1.5   \\
    &$0.2e-4$ &29.91 & 0.9184 & 0.1866 & 2.53  & 13.92  &  & 1.4  & 1.7   \\ \hline
    
    \multirow{4}{*}{Kitchen} 
    &$4e-4$ &31.55 & 0.9276 & 0.1279 & 7.67  & 7.67  & \multirow{4}{*}{4206} & 8.0  & 12.3   \\
    &$2.5e-4$ &31.78 & 0.9303 & 0.1249 & 2.65  & 10.32  &  & 1.5  & 1.7   \\
    &$1e-4$ &31.83 & 0.9309 & 0.1240 & 2.90  & 13.22  &  & 1.6  & 1.9   \\
    &$0.2e-4$ &31.83 & 0.9311 & 0.1236 & 3.03  & 16.25  &  & 1.7  & 2.0   \\   \hline

    \multirow{4}{*}{Bonsai} 
    &$4e-4$ &33.20 & 0.9484 & 0.1824 & 7.44  & 7.44  & \multirow{4}{*}{3477} & 8.3  & 13.0   \\
    &$2.5e-4$ &33.42 & 0.9505 & 0.1797 & 2.61  & 10.05  &  & 1.6  & 1.8   \\
    &$1e-4$ &33.48 & 0.9509 & 0.1791 & 3.03  & 13.08  &  & 1.7  & 2.0   \\
    &$0.2e-4$ &33.49 & 0.9510 & 0.1788 & 3.33  & 16.41  &  & 1.9  & 2.4   \\  \hline

    \multirow{4}{*}{Flowers} 

    &$4e-4$ &21.33 & 0.5744 & 0.3806 & 15.73  & 15.73  & \multirow{4}{*}{3847} & 17.2  & 28.2   \\
    &$2.5e-4$ &21.36 & 0.5762 & 0.3792 & 5.09  & 20.82  &  & 3.1  & 3.3   \\
    &$1e-4$ &21.36 & 0.5767 & 0.3788 & 5.77  & 26.58  &  & 3.3  & 3.7   \\
    &$0.2e-4$ &21.36 & 0.5769 & 0.3785 & 6.32  & 32.91  &  & 3.7  & 4.5   \\  \hline

    \multirow{4}{*}{Treehill} 
    &$4e-4$ &23.32 & 0.6416 & 0.3682 & 16.24  & 16.24  & \multirow{4}{*}{4008} & 18.2  & 29.2   \\
    &$2.5e-4$ &23.35 & 0.6431 & 0.3664 & 5.47  & 21.71  &  & 3.3  & 3.5   \\
    &$1e-4$ &23.35 & 0.6436 & 0.3655 & 6.08  & 27.79  &  & 3.4  & 3.8   \\
    &$0.2e-4$ &23.35 & 0.6439 & 0.3649 & 6.57  & 34.37  &  & 3.8  & 4.5   \\  \hline
    \multirow{4}{*}{\textbf{Avg.}}
    &$4e-4$&27.81 & 0.8079 & 0.2404 & 12.19  & 12.19  &\multirow{4}{*}{4044}  & 13.0  & 20.5   \\
    &$2.5e-4$&27.93 & 0.8103 & 0.2377 & 3.95  & 16.14  &  & 2.4  & 2.6   \\
    &$1e-4$&27.95 & 0.8109 & 0.2366 & 4.42  & 20.56  &  & 2.5  & 2.8   \\
    &$0.2e-4$&27.96 & 0.8111 & 0.2361 & 4.79  & 25.34  &  & 2.8  & 3.4   \\
    
    \toprule[2pt]
    \end{tabular}
    \vspace{-8pt}
    \caption{
    Our results of each scene on the \textbf{Mip-NeRF360} dataset~\cite{mip360} trained with $40k$ iterations. Times are measured in Second (s) and sizes are measured in MegaByte (MB). The $\Delta$ Size represents the incremental size of each level. 
    }
    \label{tab:mip360}
\end{table*}

\begin{table*}[ht]
\small
\centering
    
    \vspace{-6pt}
    \begin{tabular}{c|c|ccccc|c|cc}
    \toprule[2pt]
    Scenes  & $\lambda_s$ & PSNR$\uparrow$ & SSIM$\uparrow$ & LPSPI$\downarrow$ &$\Delta$ Size$\downarrow$ &Size$\downarrow$ &Train time &Enc  time &Dec time \\ \toprule
    \multirow{3}{*}{Chair} 
    &$4e-4$ &34.61 & 0.9833 & 0.0157 & 1.23 & 1.23  & \multirow{3}{*}{1223} & 1.1 & 1.4   \\
    &$2e-4$ &35.29 & 0.9856 & 0.0141 & 0.35 & 1.57 &  & 0.3 & 0.3   \\
    &$0.25e-4$ &35.45 & 0.9861 & 0.0136 & 0.47 & 2.05 &  & 0.3 & 0.4   \\  \hline

    \multirow{3}{*}{Drums}
    &$4e-4$ &26.31 & 0.9504 & 0.0424 & 1.68 & 1.68  & \multirow{3}{*}{1225} & 1.4 & 2.0   \\
    &$2e-4$ &26.47 & 0.9522 & 0.0407 & 0.47 & 2.15 &  & 0.3 & 0.3   \\
    &$0.25e-4$ &26.49 & 0.9524 & 0.0405 & 0.54 & 2.69 &  & 0.3 & 0.4   \\  \hline

    \multirow{3}{*}{Ficus}
    &$4e-4$ &34.78 & 0.9844 & 0.0144 & 1.18 & 1.18  & \multirow{3}{*}{1191} & 0.9 & 1.2   \\
    &$2e-4$ &35.45 & 0.9864 & 0.0129 & 0.29 & 1.47 &  & 0.2 & 0.2   \\
    &$0.25e-4$ &35.53 & 0.9866 & 0.0127 & 0.35 & 1.82 &  & 0.2 & 0.3   \\     \hline

    \multirow{3}{*}{Hotdog} 
    &$4e-4$ &37.18 & 0.9817 & 0.0277 & 0.99 & 0.99  & \multirow{3}{*}{1208}& 0.7 & 0.8   \\
    &$2e-4$ &37.77 & 0.9834 & 0.0257 & 0.21 & 1.20 &  & 0.2 & 0.2   \\
    &$0.25e-4$ &37.88 & 0.9838 & 0.0250 & 0.28 & 1.48 &  & 0.2 & 0.3   \\ \hline
    
    \multirow{3}{*}{Lego} 
    &$4e-4$ &35.08 & 0.9790 & 0.0207 & 1.45 & 1.45  & \multirow{3}{*}{1177} & 1.2 & 1.7   \\
    &$2e-4$ &35.60 & 0.9811 & 0.0190 & 0.40 & 1.85 &  & 0.3 & 0.3   \\
    &$0.25e-4$ &35.70 & 0.9814 & 0.0186 & 0.50 & 2.36 &  & 0.3 & 0.4   \\ \hline
    
    \multirow{3}{*}{Materials} 
    &$4e-4$ &30.46 & 0.9604 & 0.0407 & 1.56 & 1.56  & \multirow{3}{*}{1181} & 1.2 & 1.7   \\
    &$2e-4$ &30.75 & 0.9625 & 0.0385 & 0.42 & 1.99 &  & 0.3 & 0.3   \\
    &$0.25e-4$ &30.79 & 0.9628 & 0.0382 & 0.48 & 2.47 &  & 0.3 & 0.3   \\   \hline

    \multirow{3}{*}{Mic} 
    &$4e-4$ &36.00 & 0.9906 & 0.0093 & 1.03 & 1.03  & \multirow{3}{*}{1151} & 0.7 & 0.9   \\
    &$2e-4$ &36.58 & 0.9917 & 0.0081 & 0.20 & 1.22 &  & 0.1 & 0.2   \\
    &$0.25e-4$ &36.76 & 0.9920 & 0.0077 & 0.29 & 1.52 &  & 0.2 & 0.3   \\  \hline

    \multirow{3}{*}{Ship} 
    &$4e-4$ &31.30 & 0.9032 & 0.1170 & 2.35 & 2.35  & \multirow{3}{*}{1362} & 2.1 & 3.2   \\
    &$2e-4$ &31.53 & 0.9054 & 0.1154 & 0.72 & 3.06 &  & 0.4 & 0.5   \\
    &$0.25e-4$ &31.57 & 0.9057 & 0.1149 & 0.80 & 3.87 &  & 0.5 & 0.6   \\  \hline
    
    \multirow{3}{*}{\textbf{Avg.}}
    &$4e-4$&33.21 & 0.9666 & 0.0360 & 1.43 & 1.43 & \multirow{3}{*}{1215}  & 1.2 & 1.6   \\
    &$2e-4$&33.68 & 0.9685 & 0.0343 & 0.38 & 1.82 &  & 0.2 & 0.3   \\
    &$0.25e-4$&33.77 & 0.9689 & 0.0339 & 0.46 & 2.28 &  & 0.3 & 0.4   \\
    
    \toprule[2pt]
    \end{tabular}
    \vspace{-8pt}
    \caption{
    Our results of each scene on the \textbf{Synthetic-NeRF} dataset~\cite{NeRF} trained with $40k$ iterations. Times are measured in Second (s) and sizes are measured in MegaByte (MB). The $\Delta$ Size represents the incremental size of each level.
    }
    \label{tab:nerf}
\end{table*}

\begin{table*}[ht]
\begin{small}
\centering
    \vspace{-6pt}
    \begin{tabular}{c|ccccc|c|cc}
    \toprule[2pt]
    $\lambda_s$ & PSNR$\uparrow$ & SSIM$\uparrow$ & LPSPI$\downarrow$ &$\Delta$ Size$\downarrow$ &Size$\downarrow$ &Train time &Enc  time &Dec time \\ \toprule

    \multicolumn{9}{c}{\textbf{Mip-NeRF360}~\cite{mip360}} \\ \hline
    $4e-4$&27.69 & 0.8082 & 0.2370 & 12.64  & 12.64  & \multirow{4}{*}{2704} & 13.5  & 21.3   \\
    $2.5e-4$&27.81 & 0.8105 & 0.2346 & 4.04  & 16.68  &  & 2.5  & 2.7   \\
    $1e-4$&27.83 & 0.8110 & 0.2340 & 4.54  & 21.22  &  & 2.6  & 2.9   \\
    $0.2e-4$&27.83 & 0.8111 & 0.2336 & 4.95  & 26.17  &  & 2.9  & 3.5   \\ \toprule[1pt]

    \multicolumn{9}{c}{\textbf{Tank\&Temples}~\cite{tant}} \\  \hline
    $8e-4$ &24.27 & 0.8501 & 0.1841 & 6.31  & 6.31  & \multirow{3}{*}{1675} & 7.2  & 10.8   \\
    $4e-4$ &24.37 & 0.8537 & 0.1803 & 2.05  & 8.36  &  & 1.4  & 1.6   \\
    $0.5e-4$ &24.39 & 0.8544 & 0.1794 & 2.62  & 10.98  &  & 1.6  & 1.9   \\ \toprule[1pt]

    \multicolumn{9}{c}{\textbf{DeepBlednding}~\cite{deepblending}} \\  \hline
    $8e-4$ &30.14 & 0.9062 & 0.2638 & 3.71  & 3.71  & \multirow{3}{*}{1670} & 4.0  & 5.6   \\
    $4e-4$ &30.32 & 0.9091 & 0.2599 & 1.26  & 4.97  &  & 0.9  & 1.0   \\
    $0.5e-4$ &30.35 & 0.9096 & 0.2591 & 1.67  & 6.63  &  & 1.0  & 1.2   \\ \toprule[1pt]

    \multicolumn{9}{c}{\textbf{BungeeNeRF}~\cite{BungeeNeRF}} \\  \hline
    $8e-4$ &27.01 & 0.8730 & 0.2062 & 14.07  & 14.07  & \multirow{3}{*}{2807} & 14.3  & 20.1   \\
    $4e-4$ &27.22 & 0.8786 & 0.2004 & 4.76  & 18.83  &  & 2.8  & 3.1   \\
    $0.5e-4$ &27.25 & 0.8797 & 0.1985 & 5.47  & 24.30  &  & 3.2  & 3.8   \\ \toprule[1pt]

    \multicolumn{9}{c}{\textbf{Synthetic-NeRF}~\cite{NeRF}} \\  \hline
    $4e-4$ &33.04 & 0.9660 & 0.0365 & 1.48  & 1.48  & \multirow{3}{*}{843} & 1.3  & 1.8   \\
    $2e-4$ &33.57 & 0.9683 & 0.0344 & 0.39  & 1.88  &  & 0.3  & 0.3   \\
    $0.25e-4$ &33.66 & 0.9686 & 0.0340 & 0.48  & 2.35  &  & 0.4  & 0.4   \\

    \toprule[2pt]
    \end{tabular}
    \vspace{-8pt}
    \caption{Our results of the per-scene \textbf{averaged} results on various datasets trained with $30k$ iterations. Times are measured in Second (s) and sizes are measured in MegaByte (MB). The $\Delta$ Size represents the incremental size of each level.}
    \label{tab:30k_results}
\end{small}
\end{table*}

\begin{table*}[ht]
    \centering
    
    \small
    \vspace{-10pt}
        \begin{tabular}{l|ll|cccc|cccc}
        \toprule
        &\multicolumn{2}{l|}{\textbf{Datasets}}          & \multicolumn{4}{c|}{\textbf{Mip-NeRF360~\cite{mip360}}} & \multicolumn{4}{c}{\textbf{Tank\&Temples~\cite{tant}}} \\
        &\multicolumn{2}{l|}{\textbf{methods}} & PSNR$\uparrow$    & SSIM$\uparrow$   & LPIPS$\downarrow$ & Size$\downarrow$   & PSNR$\uparrow$   & SSIM$\uparrow$   & LPIPS$\downarrow$ & Size$\downarrow$   \\
        \bottomrule
        \multirow{2}{*}{\makecell{Base\\models}} &\multicolumn{2}{l|}{\textbf{3DGS~\cite{3DGS}}}&27.46&{0.812}&{0.222} &750.9&23.69&0.844&0.178  &431.0    \\
        &\multicolumn{2}{l|}{\textbf{Scaffold-GS~\cite{scaffold}}}&27.50&0.806&0.252  &253.9&23.96&0.853&{0.177} &86.50    \\  \hline
        
        \multirow{17}{*}{\makecell{Single\\-rate\\methods}}
        &\multicolumn{2}{l|}{\textbf{Lee~\etal~\cite{Lee}}}&27.08&0.798&0.247 &48.80&23.32&0.831&0.201&39.43    \\  
        &\multicolumn{2}{l|}{\textbf{Compressed3D~\cite{Simon}}}&3.68&26.98&0.801  &0.238&28.80&23.32&0.832&0.194    \\  
        &\multicolumn{2}{l|}{\textbf{EAGLES\cite{EAGLES}}}&27.14&0.809&0.231 &58.91&23.28&0.835&0.203&28.99    \\
        &\multicolumn{2}{l|}{\textbf{LightGaussian~\cite{Lightgaussian}}}&27.00&0.799&0.249 &44.54&22.83&0.822&0.242&22.43    \\  
        &\multicolumn{2}{l|}{\textbf{SOG~\cite{SOG}}}&26.56&0.791&0.241 &16.70&23.15&0.828&0.198&9.30    \\
        &\multicolumn{2}{l|}{\textbf{Navaneet~\etal~\cite{Compact3d}}}&27.12&0.806&0.240 &19.33&23.44&0.838&0.198&12.50    \\
        &\multicolumn{2}{l|}{\textbf{Reduced3DGS~\cite{papantonakis}}}&27.19&0.807&{0.230} &29.54&23.57&0.840&0.188&14.00    \\
        &\multicolumn{2}{l|}{\textbf{RDOGaussian~\cite{RDOGaussian}}}&27.05&0.802&0.239 &23.46&23.34&0.835&0.195&12.03    \\
        &\multicolumn{2}{l|}{\textbf{MesonGS-FT~\cite{MesonGS}}}&26.99&0.796&0.247 &27.16&23.32&0.837&0.193&16.99    \\  

        \cline{2-11}
        &\multirow{2}{*}{\textbf{HAC~\cite{HAC}}} &&27.53&0.807&0.238 &15.26&24.04&0.846&0.187&8.10    \\  
        & &&{27.77}&{0.811}&{0.230}&21.87&{24.40}&0.853&{0.177} &11.24    \\     \cline{2-11}
        &\multirow{2}{*}{\textbf{ContextGS~\cite{Contextgs}}} &&27.62&0.808&0.237 &12.68&24.20&0.852&0.184&7.05    \\  
        &&&27.75&{0.811}&0.231 &18.41&24.29&{0.855}&{0.176} &11.80    \\     \cline{2-11}
        &\multirow{2}{*}{\textbf{CompGS~\cite{liu2024compgs}}} &&26.37&0.778&0.276 &{8.83}&23.11&0.815&0.236&{5.89}    \\
        &&&27.26&0.803&0.239  &16.50&23.70&0.837&0.208&9.60    \\       \cline{2-11}
        &\multirow{2}{*}{\textbf{CAT-3DGS~\cite{CAT3DGS}}} 
        && 25.82 &0.730 &0.362  &1.72 &22.97 &0.786 &0.293 &1.42      \\
        &&& 27.77 &0.809 &0.241 &12.35 &24.41 &0.853 &0.189 &6.93 \\  \cline{2-11}
         &\multirow{2}{*}{\textbf{HEMGS~\cite{HEMGS}}} 
         &&27.74 &0.807 &0.249 &11.96 &24.40 &0.848 &0.192 &5.75 \\
         &&&27.94 &0.813 &0.230 &20.03 &24.55 &0.856 &0.176 &9.67 \\  \cline{2-11}
        &\multirow{2}{*}{\textbf{HAC++~\cite{HAC++}}} &&27.60&0.803&0.253  &{8.34} &24.22&0.849&0.190&{5.18}    \\  
        &&&{27.82}&{0.811}&0.231 &18.48&{24.32}&{0.854}&0.178 &8.63    \\     \hline
        \multirow{4}{*}{\makecell{Progre\\-ssive\\methods}}
        &\multirow{2}{*}{\textbf{GoDe \cite{gode}}} 
        &&27.07 &0.780 &0.336 &7.5 &23.72 &0.830 &0.258 &5.9 \\
        &&& 27.75 &0.810 &0.284  &19.7  &24.01 &0.846 &0.226  &11.0 \\ \cline{2-11}
        &\multirow{2}{*}{\textbf{Our PCGS}} 
        &&27.82	&0.808	&0.240	 &12.05   &24.501	&0.852	&0.186	 &5.67 \\
        &&&27.96	&0.811	&0.236	 &24.74   &24.59	&0.856	&0.181	&9.87 \\
        \toprule
    \end{tabular}

    \centering
    \begin{tabular}{l|ll|cccc|cccc}
        \toprule
        &\multicolumn{2}{l|}{\textbf{Datasets}} & \multicolumn{4}{c|}{\textbf{DeepBlending~\cite{deepblending}}} & \multicolumn{4}{c}{\textbf{BungeeNeRF~\cite{BungeeNeRF}}} \\
        &\multicolumn{2}{l|}{\textbf{methods}} & PSNR$\uparrow$    & SSIM$\uparrow$   & LPIPS$\downarrow$ & Size$\downarrow$ & PSNR$\uparrow$    & SSIM$\uparrow$   & LPIPS$\downarrow$ & Size$\downarrow$    \\
        \bottomrule
        \multirow{2}{*}{\makecell{Base\\models}}
        &\multicolumn{2}{l|}{\textbf{3DGS~\cite{3DGS}}}&29.42&0.899&{0.247} &663.9&24.87&0.841&0.205&1616    \\
        &\multicolumn{2}{l|}{\textbf{Scaffold-GS~\cite{scaffold}}}&30.21&0.906&0.254 &66.00 &26.62&0.865&0.241&183.0   \\  \hline

        \multirow{17}{*}{\makecell{Single\\-rate\\methods}}
        &\multicolumn{2}{l|}{\textbf{Lee~\etal~\cite{Lee}}}&29.79&0.901&0.258 &43.21  &23.36&0.788&0.251&82.60 \\  
        &\multicolumn{2}{l|}{\textbf{Compressed3D~\cite{Simon}}}&29.38&0.898&0.253 &25.30 &24.13&0.802&0.245&55.79   \\  
        &\multicolumn{2}{l|}{\textbf{EAGLES\cite{EAGLES}}}&29.72&0.906&{0.249} &52.34 &25.89&0.865&{0.197}&115.2  \\
        &\multicolumn{2}{l|}{\textbf{LightGaussian~\cite{Lightgaussian}}}&27.01&0.872&0.308 &33.94  &24.52&0.825&0.255&87.28  \\  
        &\multicolumn{2}{l|}{\textbf{SOG~\cite{SOG}}}&29.12&0.892&0.270 &5.70 &22.43&0.708&0.339&48.25   \\
        &\multicolumn{2}{l|}{\textbf{Navaneet~\etal~\cite{Compact3d}}}&29.90&0.907&0.251 &13.50 &24.70&0.815&0.266&33.39    \\
        &\multicolumn{2}{l|}{\textbf{Reduced3DGS~\cite{papantonakis}}}&29.63&0.902&{0.249} &18.00 &24.57&0.812&0.228&65.39   \\
        &\multicolumn{2}{l|}{\textbf{RDOGaussian~\cite{RDOGaussian}}}&29.63&0.902&0.252 &18.00 &23.37&0.762&0.286&39.06  \\
        &\multicolumn{2}{l|}{\textbf{MesonGS-FT~\cite{MesonGS}}}&29.51&0.901&0.251 &24.76 &23.06&0.771&0.235&63.11   \\    \cline{2-11}

        
        &\multirow{2}{*}{\textbf{HAC~\cite{HAC}}} &&29.98&0.902&0.269 &4.35  &26.48&0.845&0.250&18.49   \\  
        &&&{30.34}&0.906&0.258 &6.35  &27.08&0.872&0.209&29.72    \\    \cline{2-11}
        &\multirow{2}{*}{\textbf{ContextGS~\cite{Contextgs}}} &&30.11&0.907&0.265 &{3.45} &26.90&0.866&0.222&{14.00}    \\  
        &&&{30.39}&{0.909}&0.258 &6.60 &{27.15}&{0.875}&0.205&21.80   \\   \cline{2-11}
        &\multirow{2}{*}{\textbf{CompGS~\cite{liu2024compgs}}} &&29.30&0.895&0.293 &6.03  &/&/&/&/  \\
        &&&29.69&0.901&0.279 &8.77  &/&/&/&/  \\    \cline{2-11}

        &\multirow{2}{*}{\textbf{CAT-3DGS~\cite{CAT3DGS}}} 
        &&28.53 &0.878 &0.336  &0.93 &25.19 &0.808 &0.279  &10.14  \\
        &&&30.29 &0.909 &0.269  &3.56 &27.35 &0.886 &0.183  &26.59 \\    \cline{2-11}
        &\multirow{2}{*}{\textbf{HEMGS~\cite{HEMGS}}} 
        &&30.24 &0.909 &0.270  &2.85 &/&/&/&/   \\
        &&&30.40 &0.912 &0.258  &6.39 &/&/&/ &/  \\    \cline{2-11}
        
        &\multirow{2}{*}{\textbf{HAC++~\cite{HAC++}}} &&30.16&0.907&0.266 &{2.91} &26.78&0.858&0.235&{11.75}   \\  
        &&&{30.34}&{0.911}&0.254 &5.28  &{27.17}&{0.879}&{0.196}&20.82 \\ \hline

        \multirow{4}{*}{\makecell{Progressive\\methods}}
        &\multirow{2}{*}{\textbf{GoDe \cite{gode}}} 
        &&29.76 &0.897 &0.358 &3.9  &/&/&/&/   \\
        &&&30.34 &0.909 &0.331 &8.4  &/&/&/&/   \\ \cline{2-11}
        &\multirow{2}{*}{\textbf{Our PCGS}} 
        &&30.19	&0.907	&0.264	 &3.26 &26.89	&0.864	&0.223	&13.0 \\
        &&&30.38	&0.910	&0.259	 &5.81 &27.15	&0.872	&0.213	&22.8\\

        \toprule
    \end{tabular}
    
    \vspace{-10pt}
    \caption{Quantitative results of comparison methods. Sizes are measured in MegaByte (MB).
    }
    \label{tab:main_quantitative}
    
    \vspace{-15pt}
\end{table*}

\end{document}